# A Column Streaming-Based Convolution Engine and Mapping Algorithm for CNN-based Edge AI accelerators

Weison Lin and Tughrul Arslan
Institute for Integrated Micro and Nano Systems
School of Engineering, The University of Edinburgh
Edinburgh, UK
Email: {Weison.Lin, Tughrul.Arslan }@ed.ac.uk

*Abstract*—Edge AI accelerators have been emerging as a solution for near customers' applications in areas such as unmanned aerial vehicles (UAVs), image recognition sensors, wearable devices, robotics, and remote sensing satellites. These applications not only require meeting performance targets but also meeting strict area and power constraints due to their portable mobility feature and limited power sources. As a result, a column streaming-based convolution engine has been proposed in this paper that includes column sets of processing elements design for flexibility in terms of the applicability for different CNN algorithms in edge AI accelerators. Comparing to a commercialized CNN accelerator, the key results reveal that the column streaming-based convolution engine requires similar execution cycles for processing a 227 x 227 feature map with avoiding zero-padding penalties.

## I. Introduction

Convolution neural network (CNN), which has been applied to image recognition, is a kind of machine learning algorithm. CNN is usually adopted by software programs that are supported by the Artificial intelligence (AI) framework, such as TensorFlow and Caffe. These programs are usually run by central processing units (CPUs) or graphics processing units (GPUs) to form the AI systems which construct the image recognition models. The models which are trained by massive data such as big data and infer the result by the given data have been commonly seen running on cloud-based systems.

Hardware platforms for running AI technology can be sorted into the following hierarchies: data center bound system, edge-cloud coordination system, and 'edge' AI devices. The three hierarchies, computation ability, area size, and power consumption of hardware platforms from the data center to edge devices require different hardware resources and are exploited by various applications according to their demands. The state-of-the-art applications for image recognition such as unmanned aerial vehicles (UAVs), image recognition sensors, wearable devices, robotics, and remote sensing satellites belong to the third hierarchy are called edge devices. Edge devices refer to the devices connecting to the internet but near the consumers or at the edge of the whole Internet of things (IoT) system. This indicates the size of the edge devices is limited. They are also called edge AI devices when they utilize AI algorithms. The AI algorithm targeted in this paper is CNN.

The most important feature of these edge AI devices is the real-time computing ability for predicting or inferring the next decision by pre-trained data. For practicing CNN algorithms, CPUs and GPUs have been used a lot in the first two hierarchies of AI hardware platforms. Due to the inflexibility of CPUs and the high power consumption of GPUs, they are not suitable for power-sensitive edge AI devices. As a result, edge AI devices with limited power resources such as batteries require a new customized and flexible AI hardware platform to implement arbitrary CNN algorithms for real-time computing with low power consumption. To provide a flexible platform for edge AI accelerator targeting CNN algorithm, this paper proposed a column streaming-based convolution engine, inspired by [1] and [2]. The column streaming-based convolution engine does not only give out an efficient data stream but also provides configurable connections for PEs. Besides, in some filter-size cases, there are unused spare PEs. These spare PEs in the convolution engine are provided for flexibility to adapt arbitrary CNN algorithms.

The rest of this paper is organized as follows: Section 2 introduces the related works. Section 3 introduces the convolution neural network layers. Section 4 shows the column streaming-based convolution engine design and the key result. This section also contains the CNN mapping algorithm and shows the key components of the convolution engine. Conclusion and future works are summarized in section 5.

## II. Related works

For achieving the performance of AI algorithms, there are several design trends of a complete platform for AI systems, such as cloud training and inference, edge-cloud coordination, near-memory computing, and in-memory computing [3]. Currently, AI algorithms rely on the cloud or edge-cloud coordinating platforms, such as Nvidia's GPU-based chipsets, Xilinx's Versal platform, MediaTek's NeuroPilot platform, and Apple's A13 CPU [4]. As shown in [2][5], CPU and GPU are more suitable for data-center-bound platforms because CPU's processing ability, GPU's power consumption, and their system size do not meet the demand of low-power edge devices which are strictly power-limited and size sensitive [5].

When it comes to the edge-cloud coordination system, the second hierarchy, the network connectivity is necessary because those devices cannot run in the areas where is no network coverage. Data transfer through the network has significant latency, which is not acceptable for real-time AI applications such as security and emergency response [2]. Privacy is another concern when personal data is transferred through the internet. Power-sensitive edge devices require hardware to support high-performance AI computation with minimal power consumption in real-time. As a result, designing a reconfigurable AI hardware platform allowing adapting arbitrary CNN algorithms for low-power edge devices under no internet coverage is the trend. Besides ASIC, FPGA, a kind of reconfigurable architecture, is another choice. FPGA is in the category of fine-grained reconfigurable architecture (FGRA). FGRA contains a large amount of silicon to be allocated for interconnecting the logic together. This implies that FGRA impacts the rate for reconfiguring devices in real-time due to the larger bitstreams of instructions that are needed. As a result, CGRA is a better solution for real-time reconfiguration for mapping arbitrary CNN algorithms.

For achieving compact size, low power consumption, and computation ability of edge devices, there are several architectures and methods have been proposed, including the recently released commercial edge AI accelerators [2][7]-[16] and state-of-the-art edge accelerators based on CGRA [17]-[27]. For the accelerators releasing completed chip size, they are organized in Table I and II, which show the three key features with the evaluation of these prior arts. The three key features are computation ability, power consumption, and area size. The evaluation of the prior arts can be calculated as shown in (1). *CFixed16* represents the converted accelerator's performance in 16-bit fixed precision [28]. *p* and *s* represent power consumption $p$ $(w)$ and chip size $s$ ($mm^2$), respectively.

$$Evaluation\ value\ (E) = cFixed16 / (p \times s) \qquad (1)$$

### A. Architecture analysis and design direction

Fig. 2 shows the normalized three key features of the accelerators whose area size is smaller than 10 mm$^2$ in Table 1 and Table 2. The three key features of the accelerators are normalized to the same grade of computation ability by scaling up [17][18][23][27] and scaling down [20][22], linearly [29]. The result shows that except for [23][27], the remaining five accelerators have a similar trend in power consumption and area size. After normalization, the result emphasizes the insufficient performance of [23] and [27] for power-sensitive edge AI devices. [23] consumes too much power while [27] has a too big area size compared to its computation ability. However, if the targeting application requires ultra-low power consumption and can accept hundred-grade MOPs, [27] is a good choice. As a result, if we want to design an architecture for an edge AI accelerator in an ultra-small area (units' mm$^2$ area size), the power consumption and operation ability will be in the order of hundreds of mWs and GOPs, respectively. As shown, [2], implemented by steaming architecture, falls in the standard.

TABLE I. Prior Art Edge AI Accelerators

| Three key features and the evaluation value | Edge AI accelerators | | | |
|---|---|---|---|---|
| | Kneron 2018 [2] | Eyeriss 2016 [7] | 1.42TOPS/W 2016 [9] | Google Edge TPU 2018 [14][15] |
| Computation ability | 152 GOPs | 84 GOPs | 64 GOPs | 4 TOPs =2 TOPs |
| Precision | 16-bit Fixed | 16-bit Fixed | 16-bit Fixed [2] | INT8 |
| Power consumption | 350mW | 278 mW | 45mW | 2W (0.5W/TOPs) |
| Size | TSMC 65nm RF 1P6M Core area 2mmx2.5mm | TSMC 65nm LP 1P9M Chip size 4.0mmx4.0mm Core area 3.5mmx3.5mm | TSMC 65nm LP 1P8M Chip size 4.0mmx4.0mm | 5.0mmx5.0mm |
| Evaluation value | 86.86(core) | 18.88 24.66 (core) | 88.88 | 40.96 |

TABLE II. Coarse-grained Cell Array Accelerators

| Three key features and the evaluation value | Coarse-grained Cell Array Accelerators | | | | | |
|---|---|---|---|---|---|---|
| | ADRES 2017 [17] | VERSAT 2016 [18] | SOFT-BRAIN 2017 [20] | SURE based REDEFINE 2016 [22] | DT-CGRA 2016 [23][24] | Hetero-genous PULP 2018 [27] |
| Computation ability | 26.4 GOPs | 7.02 GOPs | 452 GOPs | ≈201.6 GOPs | 95 GOPs | 170 MOPs |
| Precision | 32-bit FP (A9) | 32-bit Fixed | 64-bit Fixed (DianNao) | 32-bit Fixed | 16-bit Fixed | 16-bits Fixed |
| Power consumption | 115.6 mW | 44 mW | 954.4mW | 1.22W | 1.79 W | 0.44 mW |
| Size | 0.64 mm$^2$ | 0.4 mm$^2$ | 3.76 mm$^2$ | 5.7 mm$^2$ | 3.79 mm$^2$ | 0.872 mm$^2$ |
| Evaluation value | 356.84 | 398.86 | 125.96 | 29.48 | 14 | 443.08 |

### III. CONVOLUTION NEURAL NETWORK LAYERS

The convolution function is applied in the convolution layer to calculate the inner product of the input features and weight filters. This process can be understood as mapping the input or previous layer's features to the next layer according to the emphasized features.

At the beginning of the convolution computation, the output features of a 2D input image will be generated by the feature detectors (a set of filters). The output features will become the next layer's input features. Each layer can have multiple input features such as the RGB color model, so the convolution features can be realized as 3D. The convolution output data is obtained by computing the inner product of the filter weight and the part of the input feature masked by the filter. Fig. 3 shows an example of a convolution layer. In Fig. 3, *k* represents the kernel size, known as the height and width of a weight filter. *n* represents the input layer's height and width, and *m* represents the output layer's height and width. Stride (*s*) means how far a weight filter would move to the right or down direction each time if the original start point is the most top left. The relationship of *m*, *n*, *s*, *k* is shown as (2).

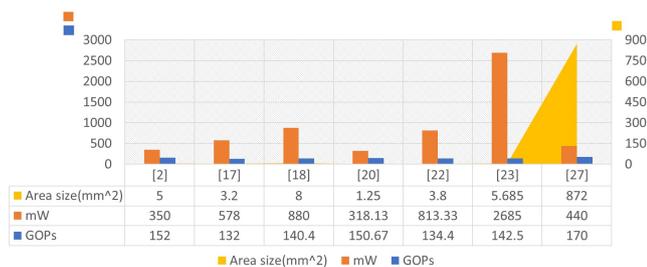

Fig. 2. Normalized to the same grade of GOPs: Three key features statistics (accelerators under 10 mm$^2$)

| | [2] | [17] | [18] | [20] | [22] | [23] | [27] |
|---|---|---|---|---|---|---|---|
| Area size(mm^2) | 5 | 3.2 | 8 | 1.25 | 3.8 | 5.685 | 872 |
| mW | 350 | 578 | 880 | 318.13 | 813.33 | 2685 | 440 |
| GOPs | 152 | 132 | 140.4 | 150.67 | 134.4 | 142.5 | 170 |

$$m = \text{floor}(\ (n-k)/s+1) \quad (2)$$

An output feature in an output layer can be obtained by the convolution computation. It can be seen as a filer scanning through an input layer according to the stride size. An example of a convolution computation in a convolution layer is shown in Fig. 4. Each output feature is obtained by the summation of the inner products, which is produced by each input feature and weight filter. After the summation is obtained, an additional bias weight (B) will be added to each summed result. How to map the convolution inputs and data flows on a hardware platform will be described in the next section.

## IV. Column Streaming Based Convolution Engine Design

The proposed convolution engine achieves reconfigurability and fewer execution cycles by using two techniques below:

1. Using filter column decomposition technique to support arbitrary kernel-sized filter's computation with column streamed input features computation method.
2. Streaming data flow to minimize bus control and using reconfigurable interconnect PE array, thus reducing hardware cost while achieving high energy efficiency.

Furthermore, in some filter-size cases, there are unused spare PEs. The PEs give out the flexibility for mapping arbitrary CNN algorithms. Although this will come with some penalties, it is always necessary to make architecture more flexible with minimum extra components.

### A. Data mapping to convolution engine

According to the survey in section II, steaming architecture is not only competitive with commercial edge AI accelerators but also CGRA edge AI accelerators.

Besides, it provides a reconfigurable streaming concept, which is flexible for adapting arbitrary CNN and avoiding unnecessary data movement. However, the reconfiguration in [2] indicates the reconfigurable weight filter composing with boundary penalty. For providing better convolution computation efficiency, this paper proposed a column streaming-based convolution engine mapping algorithm for edge AI accelerators to adapt arbitrary CNN algorithms.

Fig. 5 shows a schematic of convolution computation, which is achieved by the column streaming method. The pink square on the top-left represents the input features while the bottom-left blue square represents the weight filters. The right big blue square represents the PEs array (module), where the convolution computation is executed.

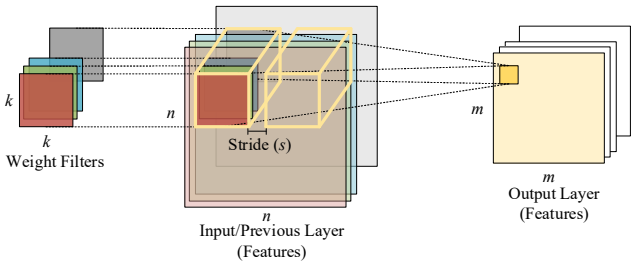

Fig. 3. Example of a convolution layer

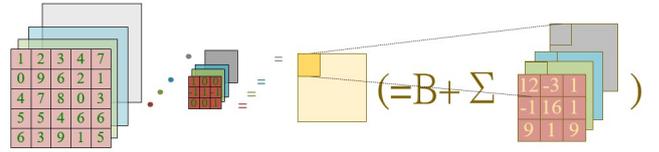

Fig. 4. Example of a convolution computation in a convolution layer

The input feature will be decomposed into a $j$-height column as the pink rectangles show in the right big blue square in Fig. 5. The weight values in the Filters block will preload to the PEs array.

Fig. 6 shows the PEs array layout in the convolution engine. The two inputs indicate the buses for the streaming input features data from memory such as SRAM. According to different filter sizes, the connections, such as wire1, wire2, wire3, wire4, will be programmed differently. Fig. 6. is an example of an input feature mapping with a 4 × 4 weight filter, so the PEs in the red squares are organized as several 4-height columns. The input lines are not fixed to be the first column and fifth column, and they will be changed depending on the mapping model, which is according to the filter size.

When the filter size is between 3 to 5, the $j$ is equal to 11. In Fig. 6, the first column set of the input feature is the $0^{th}$ data to the $20^{th}$ data of the input feature. The second column set of the input feature will be $20^{th}$ to $40^{th}$, and the $n^{th}$ set will be $(n-1)\times20$ to $n\times20$ ($n \in N_0$). The $n^{th}$ set is separated into two sub-column sets, which will be mapped to the convolution engine by bus 1 and bus 2 simultaneously in the first cycle. In the next cycle, both sub-set of the first set data will move to the right next column diagonally while the second column set will be mapped to the convolution engine by bus 1 and 2, and so on. The following sets of the input features will be streaming to the convolution engine.

For easier to understand, Fig. 6 only shows the progress view of the first 4 cycles with the first data set and some data in the second set. When the first 4 cycles are finished, data 0-22 has finished their convolution operation by doing the inner product operation with the preloaded weight values, which are in a column of the filter. Fig. 7 represents weight filters that should apply to the input features for the inner product operation. Fig. 7(a) is the weight filter adopted by Fig. 6. Fig. 8 shows that the first column of the 4 × 4 filter in Fig. 7(a) preloads to the PEs module and waits for the streaming in input features. A column of the filters' values will only be updated after a layer of feature is computed and streamed out. Then, the second column of the filter will load into the PE module and wait for the feature columns which should apply the inner product with it.

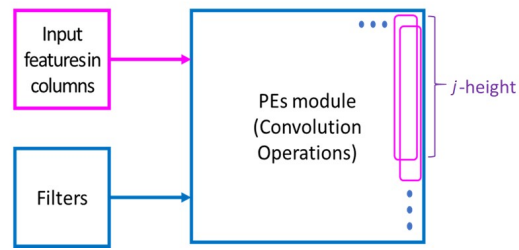

Fig. 5. Schematic of convolution computation

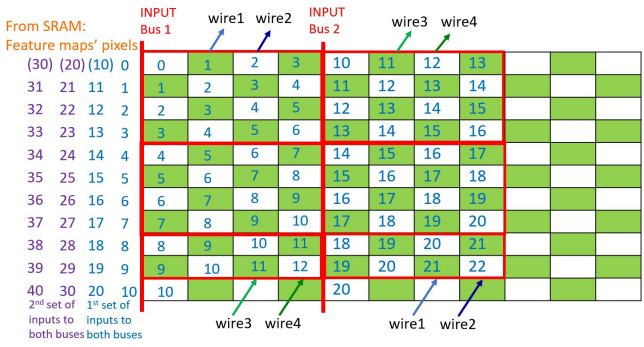

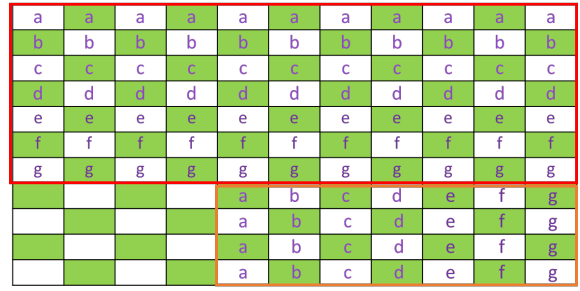

Fig. 10. The preloaded filter columns of a 7 × 7 filter in the PEs array

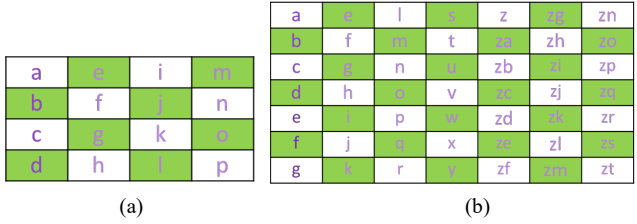

Fig. 6. Example of PEs array and the data mapping in the convolution engine (Using filter size 4 × 4 as the example @4th cycle)

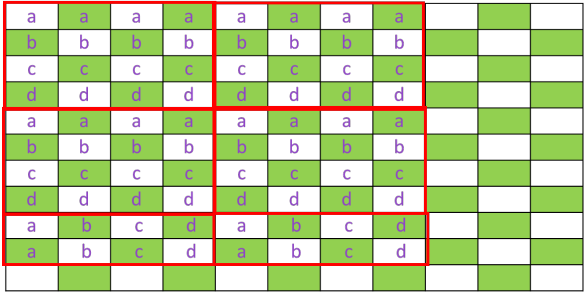

(a)  (b)

Fig. 7. (a) a 4 x 4 filter example, (b) a 7 x 7 filter example

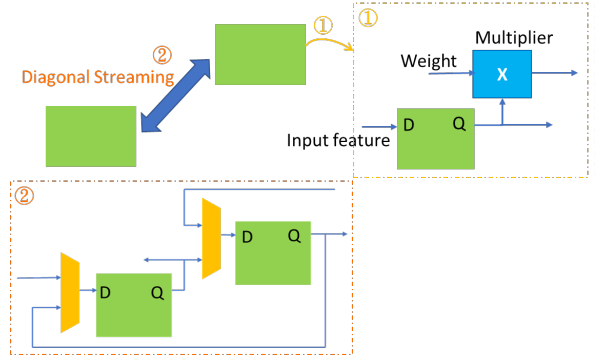

Fig. 11. Implementation of PEs in the convolution engine

Data loads into bus 1 and bus 2 will diagonally move to the next right column or left column, respectively while the next column set from memory such as SRAM will be mapped to the convolution engine by bus 1 and 2, and so on. The following sets of the input features will continue to be streaming to the convolution engine. Fig. 10 shows that the first column of the 7 × 7 filter, which is shown in Fig. 7(b), preloads to the PEs module and waits for the streaming in input features.

### B. Components of the Convolution Engine

As shown in Fig. 11 ②, the PEs are connected to the adjacent PEs diagonally. To be flexible for adapting different filters, the connections can be programmed in either direction. When Fig. 11 ① shows that the preloaded weight filter will be latched at the input of the multiplexer to wait for the streaming input features data.

### C. Key Result

For evaluating the column streaming-based convolution engine in this preliminary proposed design, we use MATLAB to count the execution cycles needed for both structures computing a 227 × 227 size feature map, which is the input size of AlexNet, one of the famous CNN.

As shown in Fig. 12, the calculation result shows that when the weight filter size is equal to 4, 7, and 10, the proposed convolution engine requires fewer cycles to compute the feature map. When the weight filter size is equal to 5 and 8, the proposed convolution engine shares similar execution cycles. When the weight filter size is equal to 3, 6, 9, and 11, the proposed convolution engine needs more cycles to compute the operations. In Fig. 12, we can observe that the orange line, indicating this work's result, goes through the blue line, [2].

Fig. 8. The preloaded filter columns of a 4 × 4 filter in the PEs array

When it comes to the filter size between 6 to 11, the streaming mapping method is different due to the filter's column containing more data. Fig. 9 is an example of an input feature mapping with a 7 × 7 weight filter, so the PEs in the red squares are organized as several 7-height columns. In Fig. 9, the first column set of the input feature is the 0[th] data to the 14[th] data of the input feature. The second column set of the input feature will be 15[th] to 29[th], and the $n$[th] set will be $(n-1)\times15$ to $n\times15-1$ ($n \in N_0$). The data 0 to 10 in the first set will load to bus 1, and the data 11-14 will load to bus 2. Except for the first set, the data, $(n-1)\times15 + i$ ($0 \leqq i \leqq 5$), in the $n$[th] set is not only loaded to bus 1 but also bus 2. The data in the $n$[th] set is one clock behind the $(n-1)$[th] set.

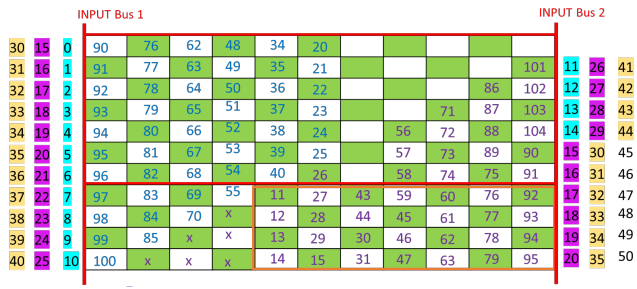

Fig. 9. Example of PEs array and the data mapping in the convolution engine (Using filter size 7 × 7 as the example @7th cycle)

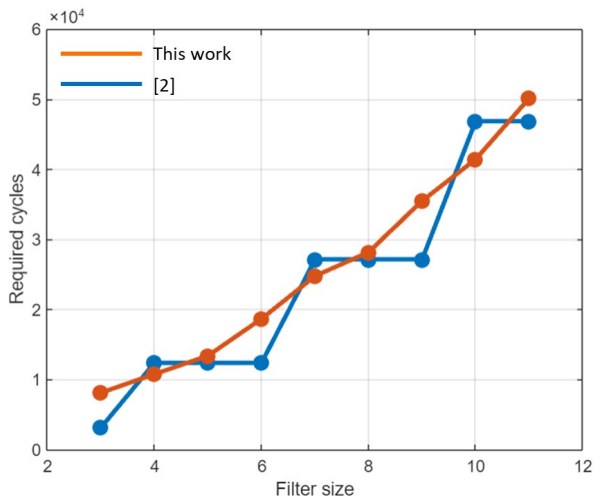

Fig. 12. Comparison: Required Cycles for a 227 × 227 feature map between this work and [2].

The orange line looks like the regression line of the blue line because we try to avoid the zero-padding boundary penalty of the [2]. Each designed working PE is all underworking without zero-padding value for certain filters.

The preliminary result shows the column streaming-based convolution engine has compatible execution ability and flexible PE design. The exact specification such as PEs amount and buses amount will be adjusted and optimized as the future work.

## V. Conclusions and Future Works

This paper has introduced a column streaming-based convolution engine for CNN with the review of up-to-date edge AI accelerators. The result reveals the proposed column streaming-based convolution engine provides compatible computation ability with similar execution cycles to [2] for processing a 227 × 227 feature map. The zero-padding boundary penalty in [2] has been avoided by using the proposed column streaming mapping algorithm.

The future work will optimize the variables such as the PE amounts and focus on the structure simulation beyond the mathematical work.

References

[1] S. Khawam, I. Nousias, M. Milward, Y. Yi, M. Muir and T. Arslan, "The Reconfigurable Instruction Cell Array," in IEEE Transactions on Very Large Scale Integration (VLSI) Systems, vol. 16, no. 1, pp. 75-85, Jan. 2008

[2] L. Du, Y. Du, Y. Li, J. Su, Y.-C. Kua, and C.-C. Liu et al., 'A Reconfigurable Streaming Deep Convolutional Neural Network Accelerator for Internet of Things', IEEE Transactions on Circuits and Systems, vol.65, no. 1, pp. 198-208, 2018.

[3] Z. You, S. Wei, H. Wu, N. Deng, M.-F. Chang, and An Chen et al., 'White paper on ai chip technologies (2018)', Tsinghua University and Beijing Innovation Centre for Future Chips.

[4] A. Montaqim, 'Top 25 ai chip companies: A macro step change inferred from the micro scale, Robotics and Automation News (2019)', May 24, URL: https://roboticsandautomationnews.com/2019/05/24/top-25-ai-chip-companies-a-macro-step-change-on-the-micro-scale/22704/

[5] K. Simonyan and A. Zisserman, 'Very deep convolution networks for large-scale image recognition', CoRR, pp.1-14, Sep. 2014.

[6] C. Clark and R. Logan, 'Power budgets for mission success (2011)', Apr. 28., unpublished. URL: http://mstl.atl.calpoly.edu/~workshop/archive/2011/Spring/Day%203/1610%20-%20Clark%20-%20Power%20Budgets%20for%20CubeSat%20Mission%20Success.pdf

[7] Y.-H. Chen, T. Krishna, J. S. Emer, and V. Sze, 'Eyeriss: An energy-efficient reconfigurable accelerator for deep convolutional neural networks', in IEEE Int. Solid-State Circuits Conf. Dig. Tech. Papers (ISSCC), San Francisco, CA, USA, Jan./Feb. 2016, pp. 262–263.

[8] C. Zhang, P. Li, G. Sun, Y. Guan, B. Xiao, and J. Cong, 'Optimizing FPGA-based accelerator design for deep convolution neural networks', in Proc. ACM/SIGDA Int. Symp. Field-Program. Gate Arrays, 2015, pp. 161–170.

[9] J. Sim, J.-S. Park, M. Kim, D. Bae, Y. Choi, and L.-S. Kim, 'A 1.42TOPS/W deep convolution neural network recognition processor for intelligent IoE systems', in IEEE Int. Solid-State Circuits Conf. (ISSCC) Dig. Tech. Papers, Jan./Feb. 2016, pp. 264–265.

[10] N. Oh, 'Intel Announces Movidius Myriad X VPU, Featuring "Neural Compute Engine"', AnandTech, August 28, 2017, URL: https://www.anandtech.com/show/11771/intel-announces-movidius-myriad-x-vpu

[11] NVIDIA, 'JETSON NANO', URL: https://developer.nvidia.com/embedded/develop/hardware

[12] Wikipedia, 'Tegra', URL: https://en.wikipedia.org/wiki/Tegra#cite_note-103

[13] Toybrick, 'TB-RK1808M0', URL: http://t.rock-chips.com/portal.php?mod=view&aid=33

[14] Coral, 'USB Accelerator', URL: https://coral.ai/products/accelerator/

[15] DIY MAKER, 'Google Coral edge TPU', 27th, Feb., 2020, URL: https://s.fanpiece.com/SmVAxcY

[16] Texas Instruments, 'AM5729 Sitara processor', URL: https://www.ti.com/product/AM5729

[17] M. Karunaratne, A. K. Mohite, T. Mitra, and L.-S. Peh, 'HyCUBE: A CGRA with Reconfigurable Single-cycle Multi-hop Interconnect', in IEEE Proc. DAC, Austin, USA, 2017, pp. 1–6.

[18] J. D. Lopes and J. T. de Sousa, 'Versat, a Minimal Coarse-Grain Reconfigurable Array', in Springer Proc. VECPAR, Porto, Portugal, 2016, pp.174–187.

[19] R. Prasad, S. Das, K. Martin, G. Tagliavini, P. Coussy, and L. Benini et al., 'TRANSPIRE: An energy-efficient TRANSprecision floatingpoint Programmable archItectuRE', in IEEE Proc. DATE, Grenoble (virtual), Germany, 2020.

[20] T. Nowatzki, V. Gangadhar, N. Ardalani, and K. Sankaralingam, 'StreamDataflow Acceleration', in IEEE Proc. ISCA, Toronto, Canada, 2017, pp. 416–429.

[21] J. Cong, H. Huang, C. Ma, B. Xiao, and P. Zhou, 'A Fully Pipelined and Dynamically Composable Architecture of CGRA', in IEEE Proc. FCCM, Boston, USA, 2014, pp. 9–16.

[22] G. Mahale, H. Mahale, S. K. Nandy, and R. Narayan, 'REFRESH: REDEFINE for Face Recognition Using SURE Homogeneous Cores', in IEEE Transactions on Parallel and Distributed Systems, vol. 27, no. 12, pp. 3602-3616, 1 Dec. 2016.

[23] X. Fan, H. Li, W. Cao, and L. Wang, 'DT-CGRA: Dual-Track CoarseGrained Reconfigurable Architecture for Stream Applications', in IEEE Proc. FPL, Lausanne, Switzerland, 2016, pp. 1–9.

[24] X. Fan, D. Wu, W. Cao, W. Luk, and L. Wang, 'Stream Processing DualTrack CGRA for Object Inference', IEEE Trans. VLSI Syst., vol. 26, no. 6, pp. 1098–1111, 2018.

[25] J. Lopes, D. Sousa, and J. C. Ferreira, 'Evaluation of CGRA architecture for real-time processing of biological signals on wearable devices', in IEEE Proc. ReConFig, Cancun, Mexico, 2017, pp. 1–7.

[26] Y.-H. Chen, T.-J. Yang, J. Emer, and V. Sze, 'Eyeriss v2: A Flexible Accelerator for Emerging Deep Neural Networks on Mobile Devices', IEEE Trans. Emerg. Sel. Topics Circuits Syst., vol. 9, no. 2, pp. 292–308, 2019.

[27] S. Das, K. J. Martin, P. Coussy, and D. Rossi, 'A Heterogeneous Cluster with Reconfigurable Accelerator for Energy Efficient Near-Sensor Data Analytics', in IEEE Proc. ISCAS, Florence, Italy, 2018, pp. 1–5.

[28] W. Lin, A. Adetomi, T. Arslan, 'Low-power ultra-small Edge AI Accelerators for Image Recognition with Convolution Neural Networks: Analysis and Future Directions', Preprints 2021, 2021070375 (doi: 10.20944/preprints202107.0375.v1).

[29] I. Takouna, W. Dawoud, and C. Meinel, 'Accurate Mutlicore Processor Power Models for Power-Aware Resource Management',